**Integrating knowledge-guided symbolic regression and model-based design of experiments to automate process flow diagram development**

Alexander W. Rogers[1], Amanda Lane[2], Cesar Mendoza[2], Simon Watson[2], Adam Kowalski[2], Philip Martin[1], Dongda Zhang[1*]

1: Department of Chemical Engineering, The University of Manchester, Oxford Road, Manchester, M1 3AL, UK.

2: Unilever R&D Port Sunlight, Quarry Road East, Bebington, CH63 3JW, UK.

*: Corresponding author: dongda.zhang@manchester.ac.uk, tel: 44 (0)161 306 5153 (Dongda Zhang).

**Abstract**

New products must be formulated rapidly to succeed in the global formulated product market; however, key product indicators (KPIs) can be complex, poorly understood functions of the chemical composition and processing history. Consequently, scale-up must currently undergo expensive trial-and-error campaigns. To accelerate process flow diagram (PFD) optimisation and knowledge discovery, this work proposed a novel digital framework to automatically quantify process mechanisms by integrating symbolic regression (SR) within model-based design of experiments (MBDoE). Each iteration, SR proposed a Pareto front of interpretable mechanistic expressions, and then MBDoE designed a new experiment to discriminate between them while balancing PFD optimisation. To investigate the framework's performance, a new process model capable of simulating general formulated product synthesis was constructed to generate in-silico data for different case studies. The framework could effectively discover ground-truth process mechanisms within a few iterations, indicating its great potential for use within the general chemical industry for digital manufacturing and product innovation.

**Keywords:** knowledge discovery, symbolic regression, model-based design of experiments, interpretable machine learning, process flow diagram optimisation.

# 1. Introduction

The global formulated products industry is large but competitive and dynamic, requiring rapid development of new products. However, the final product properties are often complex, poorly understood functions of the chemical composition and the history of processing conditions during manufacture. Hence, new product development and scale-up must undergo expensive trial-and-error campaigns that do not guarantee economic or environmental process optimality.

At this moment, model-based design of experiments (MBDoE) is the most promising approach to solving this challenge, whereby a model is used to guide exploration vs. exploitation of the experimental design space efficiently. The general MBDoE framework is flexible. The model used within MBDoE can be a mechanistic (Asprion et al., 2020; J. Taylor et al., 2021), machine learning (Castaldello et al., 2023; Gui et al., 2023) or hybrid model (Mahanty, 2023; Rogers et al., 2023). Experiments can be designed to yield the most new statistical information for the minimum amount of time and resources (Franceschini & Macchietto, 2008). If formulated as a multiobjective optimisation problem, experiments can also be designed to discover new knowledge and optimise operating conditions simultaneously (Echtermeyer et al., 2017).

However, using MBDoE for process flow diagram (PFD) development within the formulation and speciality industries remains a severe challenge due to insufficient high-quality data for pure machine learning methods or quantitative descriptions of the complex formulation processes for building hybrid or pure mechanistic models. As such, we propose that the best solution is to develop a general framework for automatically discovering good interpretable mechanistic models. By their construction, analytical expressions can be inspected, debugged, and adapted by expert practitioners to incorporate prior physical knowledge to improve data efficiency or discover new physical knowledge.

In recent years, there has been a push towards parsimonious analytical expressions with the lowest complexity required to describe the main features of the data to avoid overfitting. The sparse identification of nonlinear dynamics (SINDy) algorithm (Brunton et al., 2016) promotes sparsity among a library of candidate functions to discover ordinary differential equations (ODEs). Subsequent works extended SINDy to discover partial differential equations (PDEs) applied to fluid dynamics (Rudy et al., 2017) and implicit and rational functions under noisy conditions applied to biological networks (Kaheman et al., 2020; Mangan et al., 2016). However, such algorithms rely on the dynamics having a sparse representation in a pre-defined library. This has motivated genetic algorithms for symbolic regression (SR) that explore a much larger space of expressions by selection, mutation, and crossover defined only by a set of input features and mathematical operators (Cranmer, 2023; de Franca et al., 2023; Udrescu & Tegmark, 2020) as such, SR has had success in discovering constitutive property relationships (Karakasidis et al., 2022; Milošević et al., 2022; Papastamatiou et al., 2022) and has been applied to discovering kinetic rate models for catalytic processes (Servia et al., 2023).

However, without prior knowledge to constrain the solution space, it is very challenging for SR to find accurate expressions for complex systems – even then, the identified expression may only represent a local approximation, reducing its physical interpretability. Hence, there have been some, albeit very few, attempts to incorporate prior physical knowledge into SR (Kronberger et al., 2022; Reinbold et al., 2021), so this topic remains an open challenge.

Therefore, this work proposes a novel digital modelling framework integrating SR within MBDoE to aid automatic knowledge discovery and process flow diagram (PFD) optimisation. This framework is designed to efficiently recover underlying governing equations representing the scale-independent process dynamics through an iterative procedure. To help accelerate system identification and minimise the number of experiments required, the structure of the expressions searched by SR is constrained based on prior physical knowledge.

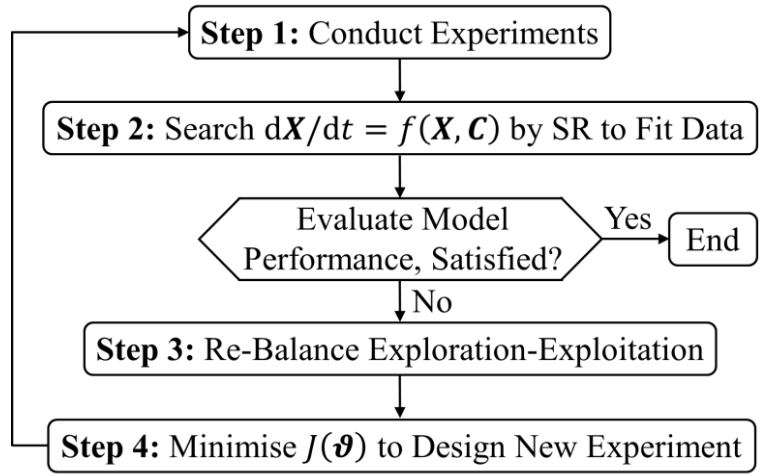

Figure 1: General SR-MBDoE flowchart for proposing mechanistic expressions for the underlying system and designing new experiments to balance exploration vs. exploitation.

The general framework integrating knowledge-guided SR and MBDoE is illustrated in Figure 1 as a flowchart. Upon starting at Step 1, an initial set of experiments is conducted based on experience or understanding of important PFD parameters, $\vartheta$, and their bounds, boxing the experimental design space. In Step 2, SR identifies a Pareto set of expressions balancing fitting accuracy and complexity for the dynamics, $dX/dt = f(X, C)$. Where $X$ and $C$ are vectors of underlying product states (e.g., in this work these are grouped phase concentrations for a multi-phase formulated product) and processing conditions (e.g., shear rate and temperature). Also observed is the KPI that is a function: $\varphi = h(X)$. The intrinsic dynamics should be independent of equipment scale and configuration, so experiments conducted at lab-scale can guide design at pilot and industrial-scale. Expressions are scored and selected to build process models. If the top scoring model or PFD performance is satisfactory, then MBDoE should be terminated, otherwise, a new experiment is conducted. In Step 4, experiments are designed by minimising the objective function, $J(\vartheta)$. The objective function, $J(\vartheta)$, can design experiments for maximum model discrimination and therefore knowledge discovery, process optimization, or balance both simultaneously. The weighting between these two objectives is automatically controlled in Step 3 based on the information and optimality improvement in the experiment from the previous MBDoE iteration.

## 2. Results

### 2.1. Mechanistic Modelling of Formulated Product Synthesis

This research used liquid products, typical of cosmetic and pharmaceutical creams as a case study for which no quantitative mechanistic models have yet been proposed to simulate these processes, presenting a severe challenge within the formulation industry for future digital manufacturing. To link the effect of phase composition, manufacturing scale and different processing variables, a new mechanistic model was proposed for the first time to approximate the product formulation. This model was used to run computational experiments and generate in-silico data to test the SR-MBDoE methodology for knowledge discovery and simultaneous PFD optimisation.

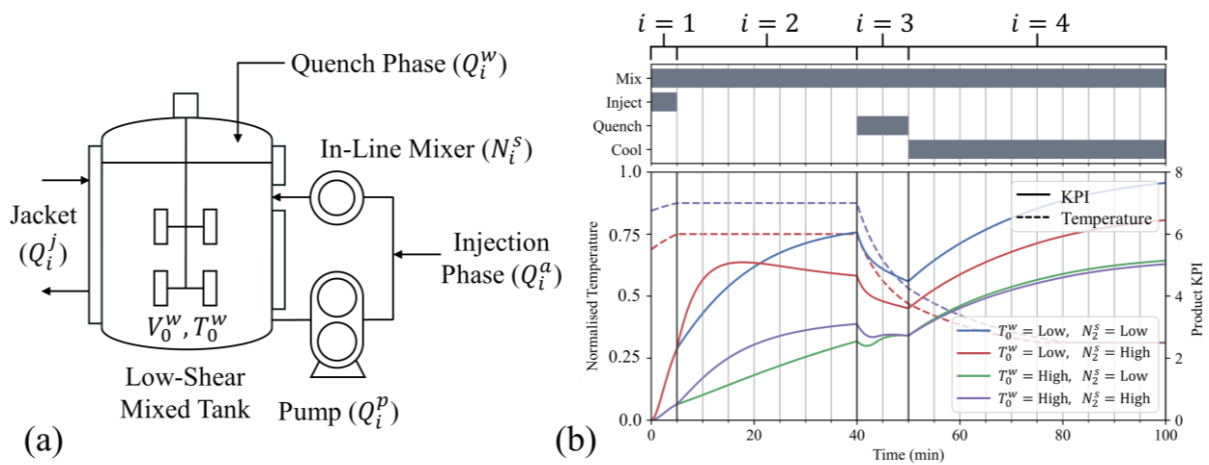

Figure 2: Equipment for formulated product manufacturing (a) where $V$ is volume, $Q$ is flowrate and $T$ is temperature; $w, a, s, p$ and $j$ denote two ingredient streams, the in-line mixer, pump, and jacket, respectively. Shown are four simulated KPI profiles from following a four-step PFD (b) where $i$ indexes the processing steps.

The synthetic model reduced the multiphase product by grouping chemical components into key phases, and mechanistic expressions (see Equations 8a, 8b and 8c in Methods) were proposed to approximate the rate of phase reaction in terms of forward and backward driving forces for each mechanism; these were the intrinsic process dynamics that SR-MBDoE aimed to rediscover. The rates, $dX/dt = f(X, C)$, were functions of the phase concentrations and intrinsic operating conditions (e.g., local temperature and average shear rate experienced by the fluid in the impeller regions). The dynamic mixture KPI, $\boldsymbol{\psi}$, was a function of the dynamic phase concentrations: $\boldsymbol{\psi} = h(X)$. These intrinsic kinetics were embedded into a mass and energy balance (see Equations 9a, 9b, 9c, 9c and 9e in Methods) for the industrially commonplace recycle emulsification configuration depicted in Figure 2a. The synthetic PFD generated from the complex mechanistic model mirrors the four steps involved in a real industrial production process (see Methods for the formulation process flow diagram). Figure 2b shows four in-silico product KPI profiles for the experiments used to initiate MBDoE in the following case studies. Note that this approach describing the dynamics of the underlying product states (i.e., in this case study, chemical constituents grouped into structural phases with measurable concentrations) by a set of differential equations and then mapping these underlying product states to product KPIs can apply generally to approximating formulated product synthesis processes.

## 2.2. Carrying Expressions Over for Incremental Improvement

To begin with, in the first case study we investigated the performance of the SR-MBDoE framework when the sole objective was knowledge discovery. Each in-silico experiment took measurements representing the grouped phase concentrations, $X$, and the local operating conditions experienced by the fluid, $C$, across the in-line mixer, where the intrinsic dynamics dominate (see Methods for in-silico experiment design and measurements). To initiate MBDoE, four synthetic experiments were conducted at high and low initial temperatures and in-line mixer speeds based on prior knowledge of important PFD parameters and their upper and lower bounds, yielding the product KPI profiles seen earlier in Figure 2b. Then, for each MBDoE iteration, SR proposed three candidate expressions describing the rates of each of the three underlying mechanisms as functions of $\boldsymbol{X}$ and $\boldsymbol{C}$ (see Methods for details on construction and selection of candidate expressions). Next, one new experiment minimising $J(\boldsymbol{\vartheta})$, that maximised the variance in the candidate's predictions for model discrimination was proposed and conducted, expanding the dataset for the next iteration.

Upon the next iteration of MBDoE with this expanded dataset, one could either propose a new set of equations from scratch or carry over the best-performing equations from the previous MBDoE iteration and make modifications to incorporate the new information. From a computational perspective, carrying over previously learnt knowledge is more efficient but risks carrying over biases because although SR by evolution is stochastic, expressions must pass through intermediate states to correct old misunderstandings. If the intermediate state is a poor-fitting candidate, the population of expressions may remain trapped in a suboptimal local solution. Thus, both approaches were investigated. Case Study 1A built new expressions from scratch, while Case Study 1B carried over the expressions from the previous MBDoE iteration.

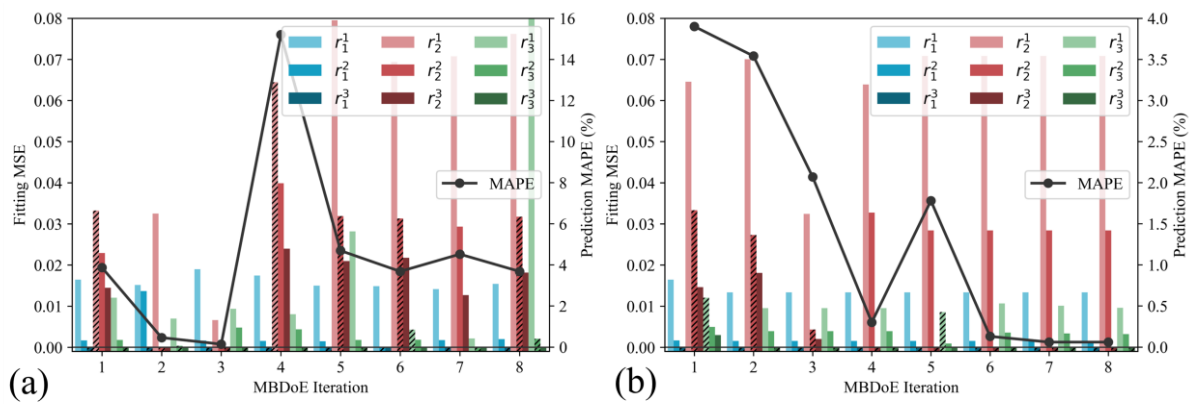

Figure 3: Mean absolute percentage error (MAPE, for prediction) and mean-square error (MSE, for fitting) for the expressions $r_i^j$ proposed at each iteration of Case Study 1A (a) and Case

Study 1B (b), where the scripts $i$ and $j$ denote the rate equation number and relative complexity. The bars corresponding to the top-scoring expressions used for prediction and MAPE estimation at each iteration are shaded black.

Figures 3a and 3b show the prediction mean absolute percentage error (MAPE) for the final product KPI and fitting mean square error (MSE) for the top three scoring expressions for each rate equation following each MBDoE iteration ($I_{\text{MBDoE}}$) for Case Studies 1A and 1B, respectively. For the MAPE to be small, the MSE for the selected expressions (i.e., the ones shaded black) for all three rate equations had at least also to be small. A detailed discussion can be found in the Supplementary, reasoning how the proposed experiment after $I_{\text{MBDoE}} = 1$ can effectively help discriminate different model structures.

Over subsequent MBDoE iterations, in Case Study 1A, the MAPE decreased from 3.87% to a minimum of 0.14% by $I_{\text{MBDoE}} = 3$, while in Case Study 1B, the MAPE decreased from 3.90% to 0.30% by $I_{\text{MBDoE}} = 4$. This demonstrates that the new information from the experiments proposed by MBDoE improved the generalisability of the expressions. To begin with, the "evolutionary pressure" towards the ground truth expressions was weak, and there were too many similarly fitting expressions to search. As more carefully designed experiments were added, the difference in the MSE between correct and incorrect expressions during tournament selection became stronger, encouraging better-fitting expressions to be promoted.

**2.3. Importance of MBDoE over Statistical Model Selection Criteria**

In this work, expressions were scored based on the negated derivative of the log-loss with respect to complexity (see Equations 1a and 1b in Methods for the model selection criterion). In Case Studies 1A and 1B, the MAPE eventually peaked in $I_{\text{MBDoE}} = 4$ and $I_{\text{MBDoE}} = 5$, respectively. In Case Study 1B, an approximation was proposed in $I_{\text{MBDoE}} = 5$ for $r_3$ that had an MSE of 0.0086, but it's much lower complexity gave it a higher score of 1.3 compared with

1.1 and 0.82 for the other two candidates with better MSEs of 0.00001 and 0.0009, respectively. In Case Study 1B, expressions were carried over to the next MBDoE iteration and the ground truth structure for $r_3$ had been discovered already but had been demoted in favour of a simpler approximation. Hence, a sufficiently simple approximation can be preferred over more complex expressions, even if that more complex expression is the ground truth. This shows how the choice of metric to score and select expressions is critical. The advantage of the score definition in Equation 1b used in this work in contrast to traditional statistical information criteria (e.g., Akaike (Akaike, 1998), Bayesian (Schwarz, 1978) and Hannan-Quinn (Hannan & Quinn, 1979)) is that expressions at 'elbow' points on the Pareto front are selected. However, other metrics will come with their own biases. Only by performing new, carefully designed experiments is it possible to reliably discriminate between different hypotheses and identify the correct model. So, when $I_{\text{MBDoE}} = 6$ added a new experiment, and the MSE of the approximation increased from 0.0086 to 0.011, the score ranking flipped, and the ground truth was correctly re-identified.

While the MAPE recovered in Case Study 1B, it did not in Case Study 1A when SR had to propose new expressions from scratch. In Case Study 1A, the MAPE remained around 4.14% for $I_{\text{MBDoE}} \geq 5$, even as a further five unique experiments were added. This is despite the fact that SR was allowed 10 times longer to search for good expressions per MBDoE iteration in Case Study 1A compared with Case Study 1B. This result suggests that building complete expressions from scratch is more challenging. So, rather than inherited biases hindering better expression discovery, adding new information by MBDoE incrementally seems to guide SR towards the correct structure more efficiently. Therefore, hereafter, expressions were carried over from one MBDoE iteration to the next to be modified using the new information.

## 2.4. Acceleration by Knowledge-Guided Symbolic Regression

Good extrapolation is key to minimising the number of experiments needed for system identification. However, without correct underlying theory with which to motivate model selection or construction, this is not guaranteed. There can be multiple possible expressions that fit the observed data well but then disagree further away, without necessarily being accurate. Information criteria (e.g., Akaike (Akaike, 1998), Bayesian (Schwarz, 1978) and Hannan-Quinn (Hannan & Quinn, 1979)) have been proposed for selecting models based on the likelihood of observing data given a certain model while penalising the number of parameters. However, these do not guarantee correct model selection. Consider fitting data points over a narrow slice of an input domain; without prior knowledge about the nature of the underlying function, a linear correlation would indeed be the simplest, well-fitting hypothesis. For this reason, physical models motivated from correct underlying theory tend to extrapolate better than statistical models that are not motivated by physical theory.

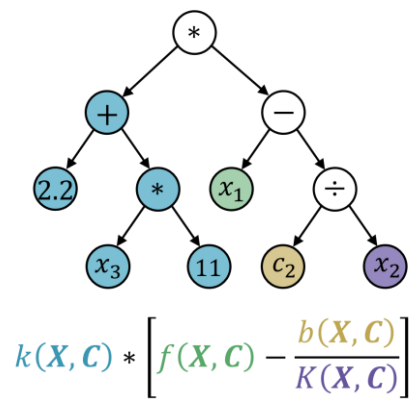

Figure 4: Expression tree constrained to a structure representing prior physical knowledge where $X$ and $C$ are vectors of intrinsic material properties and operating conditions, respectively; tree can be restricted to other forms if different prior knowledge is thought true.

Therefore, the expressions evaluated during SR were constrained to those that adhere to some physically meaningful interpretation, discarding nonconformers during genetic evolution (see Methods for implementation of structural constraints). For example, specific to the current case study, expressions were restricted to the set of structures of the form depicted in Figure 4: $k(\cdot) \times [f(\cdot) - b(\cdot) \div K(\cdot)]$, reflecting the general form of the ground truth kinetic equations in (8a), (8b) and (8c). Where $k(\cdot)$, $f(\cdot)$, $b(\cdot)$ and $K(\cdot)$ are sub-expressions representing the overall rate constants, forward and backward driving forces, and equilibrium constants, respectively, for each of the underlying formulation process mechanisms. These were found simultaneously, made partially identifiable by the cap on total complexity. Through this knowledge-guided evolution approach, one can expect SR to be more likely to discover physically insightful expressions. In addition, while the assumption of state equilibration is generally applicable to PFD optimisation, as explored in this case study, the expression tree can be restricted to other structural forms if different prior knowledge is considered to be true.

Thus far, in Case Studies 1A and 1B, this constraint (i.e., based on prior knowledge of state equilibration) has been active and successfully yielded expressions of the desired structure that could be easily interpreted in terms of key forward and backward driving force factors. Case Study 1C investigated the performance of the SR-MBDoE framework when lifting this constraint. Here, the MAPE decreased from 25.6% to 1.98% and plateaued for $I_{\text{MBDoE}} \geq 7$, demonstrating that constraining the search to the correct structure significantly improved fitting accuracy and sped up knowledge discovery when the number of experiments was small.

## 2.5. Simultaneous Knowledge Discovery and PFD Optimisation

Case Study 2 investigated the performance of the SR-MBDoE framework when the objective was simultaneous knowledge discovery and PFD optimisation. This was achieved by formulating $J(\vartheta)$ as a multiobjective optimisation problem with a weighting $0 \leq \alpha \leq 1$ determined automatically in Step 3 to re-balance the competition between exploration and exploitation based on the actual information and optimality gains from the previous MBDoE iteration (see Methods for details on the formulation of the multiobjective optimisation problem and the weighting parameter control function). In this case study, the aim was to achieve a final product KPI of $\eta_t = 7$ within a $\kappa = \pm 3\%$ tolerance, within the shortest batch time, $\tau$.

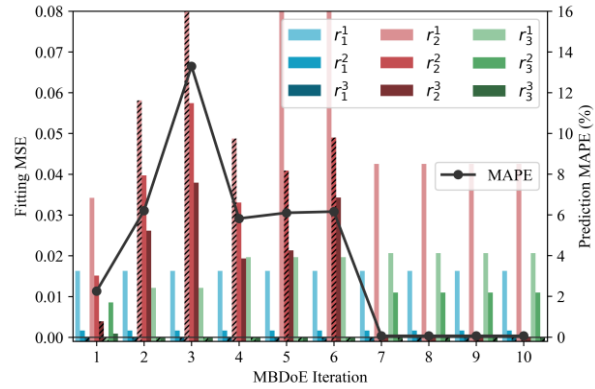

Figure 5: Mean absolute percentage error (MAPE) and mean-square error (MSE) for the expressions $r_i^j$ proposed at each iteration of Case Study 2, where the scripts $i$ and $j$ denote the rate equation number and relative complexity. Bars corresponding to the top-scoring expressions used for prediction and MAPE estimation at each iteration are shaded black.

Table 2: Maximum expected ($J_M^{max}$ and $J_P^{max}$) and actual ($\Delta J_M$ and $\Delta J_P$) information and optimality gains, respectively, and exploration-exploitation weighting ($\alpha$) and total batch time ($\tau$) for each MBDoE iteration ($I_{MBDoE}$) for Case Study 2.

| $I_{MBDoE}$ | $J_M^{max}$ | $J_P^{max}$ | $\Delta J_M$ | $\Delta J_P$ | $\alpha$ | $\tau$ (min) |
|---|---|---|---|---|---|---|
| 0 | - | - | - | - | 0.5 | 100 |
| 1 | 0.08 | 18.7 | 0.0045 | 0 | 1 | 120 |
| 2 | - | - | 0.26 | 5.4 | 0.045 | 118 |
| 3 | 0.23 | 21.2 | 1.8 | 0.078 | 0.95 | 125 |
| 4 | 0.09 | 14.2 | 0.10 | 10 | 0.010 | 107 |
| 5 | 0.50 | 14.5 | 0.30 | 1.1 | 0.20 | 69 |
| 6 | 0.46 | 14.5 | 0.041 | 0.51 | 0.074 | 144 |
| 7 | 0.15 | 19.1 | 1.3e-06 | 0.0016 | 0.00085 | 83 |
| 8 | 0.15 | 19.1 | 8.4e-07 | 0.0011 | 0.00075 | 59 |
| 9 | 0.15 | 19.1 | 7.7e-07 | 0.0011 | 0.00066 | 58 |
| 10 | 0.15 | 19.1 | 8.5e-07 | 0.0012 | 0.00070 | 57 |

Figure 5 shows the MAPE and MSE at each MBDoE iteration ($I_{\text{MBDoE}}$) for Case Study 2, while Table 2 shows the maximum expected and actual information and optimality, gains, the calculated value of the exploration-exploitation weighting, $\alpha$, and the total batch time, $\tau$, for the proposed experiment. Initially, MBDoE bounced between exploration (i.e., $\alpha > 0.5$) and exploitation (i.e., $\alpha < 0.5$). For as long as the actual information gain was large, and new experiments continued to be informative, exploration was prioritised. Once the actual information gain became small (i.e., $\alpha = 0.2$ by $I_{\text{MBDoE}} = 5$), and new experiments no longer proved to be as informative, then process optimisation was prioritised. If, at any point, the model aimed for and successfully hit on a good recipe (i.e., one that achieved an in-spec product KPI), then there would be nothing new to learn or improve about the process within the local vicinity; thus, the next iteration would bounce back to pure exploration. By $I_{\text{MBDoE}} = 7$ the ground truth had been discovered, so MBDoE switched to pure optimisation (i.e., $\alpha = 0.00085$). To more effectively terminate MBDoE in future, an MBDoE stopping criteria could be to iterate until $\alpha < \alpha_t$ drops below a threshold (e.g., $\alpha_t \approx 1 \times 10^{-3}$) signalling when the information gain is small enough to focus one final experiment on optimisation.

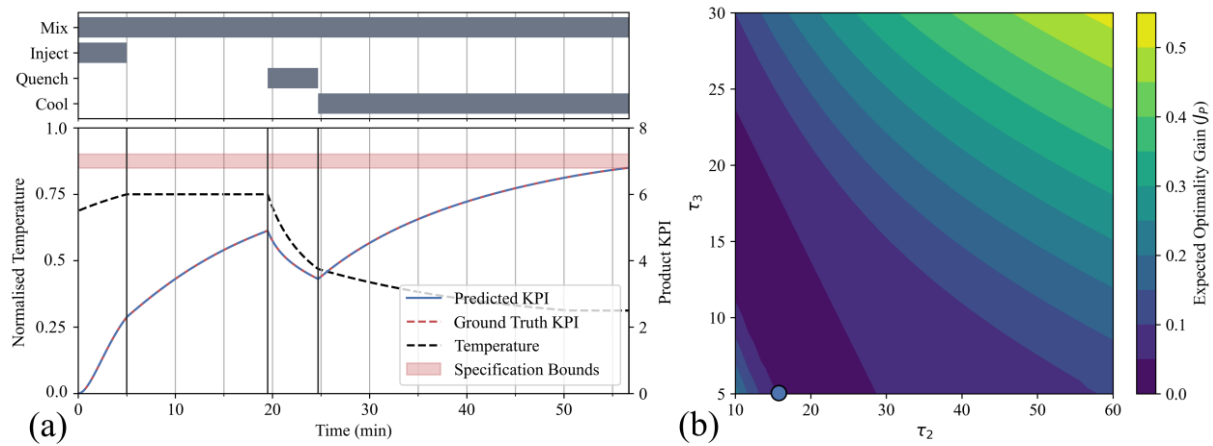

Figure 6: Predicted and actual product KPI profiles for optimised processing parameters using the model from $I_{\text{MBDoE}} = 10$ from Case Study 2 (a) and the corresponding slice of $J(\vartheta)$ about the optimum (b) for $\tau_2$ and $\tau_3$ where the blue point represents the optimised parameters. Where $\tau_2$ and $\tau_3$ are the processing step durations for steps two and three, respectively.

Figure 6a shows the trajectory for the optimal PFD designed following MBDoE, while Figure 6b shows a slice through the six-dimensional parameter space ($\boldsymbol{\vartheta}$) for the predicted optimality gain objective function. The nominal total batch time was greatly reduced from $\tau = 100$ min at $I_{\text{MBDoE}} = 0$ to $\tau = 59$ min, while the final product KPI of $\eta_f = 6.79$ was within the lower bounds of the $\kappa = 3\%$ tolerance, further evidencing the practical advantage and efficiency of the currently proposed SR-MBDoE digital framework. Due to the discovery of mechanistic rate expressions, it is possible to interpret the physical trade-offs made by the optimised PFD that would not be possible for a pure machine learning approach. A detailed discussion of the physical interpretation of the optimum PFD can be found in the Supplementary.

## 3. Discussion

Through two case studies, it was demonstrated that despite the highly complex nature of the underlying ground truth, the proposed knowledge-guided SR-MBDoE framework could recover the ground truth exactly after only a small number of experiments, demonstrating its great potential. As expected, it took more MBDoE iterations when balancing exploration and exploitation. While carrying over expressions from previous MBDoE iterations for modification proved more successful than building expressions from scratch. However, selecting expressions based on statistical parsimony alone risks bias; only by conducting carefully designed experiments was it possible to reliably discriminate between similarly fitting candidates of different complexities. Then, when the knowledge-guided constraint on the expressions' structure was lifted, the prediction accuracy for the same number of experiments decreased substantially. By synergising human intelligence with the automatic discovery and discrimination of interpretable mechanistic models representing the scale-independent process dynamics, this study demonstrates that SR-MBDoE is a highly promising approach to systematising experimental exploration for physical understanding and PFD optimisation in industries seeking to rapidly transition to more sustainable economic models or with highly competitive market conditions for continuous product or process innovation.

## 4. Methods

### 4.1. Symbolic Regression

With reference to Figure 1, Step 2 is now detailed, whereby different candidate expressions are found by SR representing the intrinsic process dynamics, subject to some prior physical knowledge. Although it allows the discovery of new mechanistic models, one of the key challenges of SR is finding good-fitting expressions efficiently. In this work, tournament selection promoted and mutated the best candidates from a population of expressions represented by directed acyclic graphs using the Python front-ended Julia library PySR by (Cranmer, 2023). In each round of tournament selection, a random subset of individuals was drawn from the population and evaluated according to a fitness function; the fittest individuals were more likely to be selected as the winner and carried forward by replacing the oldest member of the subset with a copy of the winner that has a small chance of being mutated. The fitness function is the mean square error (i.e., MSE, Equation 1a) plus the complexity of the expression multiplied by an adaptive parsimony weighting.

PySR (Cranmer, 2023) embeds the genetic algorithm inside an *evolve-simplify-optimise* loop: after a set number of tournaments and mutations, the equations are simplified using algebraic equivalencies, followed by a few iterations of local-gradient-based optimisation to refine the numerical constants in the expressions. In the end, the fittest individuals in the population at each level of complexity are lined up onto a Pareto front and scored by the negated derivative of the log-loss with respect to complexity, as shown in Equation 1b (Cranmer, 2023). In (1a)

$\mathcal{L}_i$ is the MSE between the predicted $\hat{\boldsymbol{y}}_i \in \mathbb{R}^{N \times 1}$ and measured $\boldsymbol{y} \in \mathbb{R}^{N \times 1}$ outputs weighted by the diagonal matrix $\boldsymbol{\Lambda}$ averaged over $N$ datapoints, where the subscript $i$ indexes the candidate expression from the Pareto set of expressions. Complexity, $\mathcal{C}$, is defined as the total number of operators, variables, and constants. For the ordered Pareto set: $\mathcal{C}_{i+1} > \mathcal{C}_i$.

$$\mathcal{L}_i = \frac{1}{N} \sum_{n=1}^{N} (\boldsymbol{y} - \hat{\boldsymbol{y}}_i)^{\mathrm{T}} \boldsymbol{\Lambda} (\boldsymbol{y} - \hat{\boldsymbol{y}}_i) + P_i \tag{1a}$$

$$\text{Score} = -\frac{\log(\mathcal{L}_{i+1}) - \log(\mathcal{L}_i)}{\mathcal{C}_{i+1} - \mathcal{C}_i} \tag{1b}$$

### 4.2. Constraining the Expression Structure

Expressions evaluated during SR were constrained to those that adhere to some physically meaningful interpretation, discarding nonconformers.

$$P = \begin{cases} \infty, & \text{if } G \notin k(\cdot) \times [f(\cdot) - b(\cdot) \div K(\cdot)] \\ 0, & \text{otherwise} \end{cases} \tag{2}$$

Implemented by means of the penalty, $P$, appended to (1a) defined in (2) takes infinity when the expression tree, $G$, is not in the set of desired structures. The penalty was implemented as a logical condition that examines the top-level operators (e.g., if the first operator is not multiplication: return infinity, otherwise return zero). Specific to the current case, $G$ was constrained to take the form shown in Figure 4, reflecting the general form of the kinetic equations in (8a), (8b) and (8c), constructed from forward and backward contributions to each mechanism. $k(\cdot)$, $f(\cdot)$, $b(\cdot)$ and $K(\cdot)$ are sub-expressions representing the overall rate constants, forward and backward driving forces, and equilibrium constants, respectively, for each of the underlying formulation process mechanisms.

These sub-expressions were found simultaneously, made partially identifiable by a cap on total complexity. Through this knowledge-guided evolution approach, one can expect SR to be more likely to discover physically insightful expressions. Note that while the assumption of state equilibration is generally applicable to formulation process PFD development, as explored here, the expression tree can be restricted to other structural forms if different prior knowledge is considered true. Depending on the amount of information available, the same approach can be used to constrain partially known expression structures. In the case of an infinite penalty, evaluation of $\mathcal{L}_i$ is skipped, conserving computational resources.

**4.3. General SR-MBDoE Algorithm**

With reference to Figure 1, Steps 3 and 4 are now detailed, whereby new experiments are proposed by model-based design of experiments (MBDoE) for simultaneous knowledge generation and PFD optimisation. Scientific models encode hypotheses; again, because multiple possible models will often fit the observed data well but then disagree further away, it is a fundamental part of the scientific method to design and conduct experiments to discriminate between these hypotheses. Following the flowchart in Figure 1 from Step 1, an initial set of experiments is conducted based on experience or understanding of important PFD parameters, $\boldsymbol{\vartheta}$, (e.g., ingredient additions and processing conditions) and their upper, $\boldsymbol{\vartheta}_{\text{ub}}$, and lower, $\boldsymbol{\vartheta}_{\text{lb}}$, bounds boxing the experimental design space. Then, in Step 2, SR identifies different potential expressions for the intrinsic dynamics, $d\boldsymbol{X}/dt = f(\boldsymbol{X}, \boldsymbol{C})$ balancing fitting accuracy and complexity. In Step 4, MBDoE designs new experiments by minimising the multiobjective function, $J(\boldsymbol{\vartheta})$, in Equation 3. Where $J_M(\boldsymbol{\vartheta})$ is the expected information gain and $J_P(\boldsymbol{\vartheta})$ is the objective function for process optimisation, while $J_M^{\max}$ and $J_P^{\max}$ are normalisation constants. The weighting parameter $0 \leq \alpha \leq 1$ systematically balances these two objectives and is updated in Step 3 for each new SR-MBDoE iteration.

$$\min_{\vartheta} J(\vartheta) = \alpha \cdot \frac{J_M(\vartheta)}{J_M^{\max}} + (1-\alpha) \cdot \frac{J_P(\vartheta)}{J_P^{\max}} \qquad (3a)$$

$$J_M^{\max} = \max_{\vartheta} J_M(\vartheta) \qquad (3b)$$

$$J_P^{\max} = \max_{\vartheta} J_P(\vartheta) \qquad (3c)$$

$$\text{s.t. } \vartheta_{lb} \leq \vartheta \leq \vartheta_{ub} \qquad (3d)$$

$J(\vartheta)$, $J_M(\vartheta)$ and $J_P(\vartheta)$ were each minimised in two steps: global Bayesian optimisation and then local solution refinement by the Nelder-Mead simplex algorithm, implemented using the Python libraries Optuna (Akiba et al., 2019) and SciPy (Virtanen et al., 2020), respectively. All proposed expressions were kept in symbolic form using the Python library SymPy (Meurer et al., 2017) for automatic algebraic manipulation.

### 4.4. Information and Optimality Objective Functions for PFD Development

Expected information gain, $J_M$, is designed to maximize the information of new experiments for model discrimination. The PFD parameters, $\vartheta$, prescribe the sequence of ingredient addition flowrates and processing conditions. Simulating the recipe from the start of the batch (i.e., $t = 0$) until the end of the batch (i.e., $t = \tau$) by integrating $dX/dt = f(X, C)$ a matrix of predicted concentration profiles, $\widehat{X}$, is obtained as in (4a). This is repeated using each set, $S$, of three top-scoring candidates for $dX/dt = f(X, C)$ proposed by SR. Then the KPI, $\widehat{\psi}_S$, is computed as a function of $\widehat{X}_S$, as in (4b); this can either be known, or also constructed by SR. Finally, $J_E$ is calculated as the variance in the predicted final product KPI, $\widehat{\psi}_S$, as in (4c). The superscript $k$ indexes different KPIs (e.g., physical properties) to be maximized for parallel experiments.

$$\widehat{X}_S = \int_0^\tau \frac{dX}{dt}(X, C)\bigg|_S dt \qquad (4a)$$

$$\widehat{\psi}_S = h(\widehat{X}_S) \qquad (4b)$$

$$J_E^k = -\text{var}\big([\hat{\psi}_1^k, \hat{\psi}_2^k, \dots \hat{\psi}_{N_S}^k]\big) \qquad (4c)$$

For this case study, $J_P$, was defined in Equations 5a and 5b to be the total batch time, $\tau$, plus a quadratic penalty for when the product KPI, $\hat{\psi}^k$, was not within tolerance, $\kappa$, of the target product KPI, $\psi_t^k$. Within the tolerance, only minimising the total batch time is important.

$$J_P^k = \tau + \max(\varepsilon^k, 0) \tag{5a}$$

$$\varepsilon^k = \left(\psi_t^k - \hat{\psi}^k\right)^2 - \left(\kappa \cdot \psi_t^k\right)^2 \tag{5b}$$

Given the need to consider both objective functions, it is of critical importance to balance exploration vs. exploitation automatically; Equation 6 systematically weights how much each objective was prioritised from one MBDoE iteration to the next as a function of the actual information gain, $\Delta J_M$, and optimality gain, $\Delta J_P$, based on the outcome of the experiments conducted in the previous MBDoE iteration.

$$\alpha = \frac{\Delta J_M}{\Delta J_M + \Delta J_P} \tag{6}$$

$\Delta J_M$ in Equation 7a was the error between the predicted, $\hat{\psi}^k$, and actual, $\psi_a^k$, final product KPI. $\Delta J_P$ in Equations 7b and 7c was the error between the target, $\psi_t^k$, and actual, $\psi_a^k$, final product KPI, with no further gain to be sought from hitting the target any closer than the maximum tolerance. Initially, $\alpha = 0.5$ for $I_{\text{MBDoE}} = 1$. As $\Delta J_M \ll \Delta J_P$, optimisation for an on-spec product rather than exploration is increasingly preferred.

$$\Delta J_M = \left(\psi^k - \hat{\psi}_a^k\right)^2 \tag{7a}$$

$$\Delta J_P = \max(\varepsilon_a^k, 0) \tag{7b}$$

$$\varepsilon_a^k = \left(\psi_t^k - \psi_a^k\right)^2 - \left(\kappa \cdot \psi_t^k\right)^2 \tag{7c}$$

### 4.5. Synthetic Product Model

The model reduced the multiphase product to five reacting phases $W$, $A$ $L$ $L^*$ and $V$. Three equations (8a, 8b and 8c) describe the rate of the three main mixing mechanisms. Each mechanism was a function of the concentrations of the five lumped phases $X =$

$[X_W, X_A, X_L, X_V, X_{L^*}]^T$ and processing conditions $\boldsymbol{C} = [T, \dot{\gamma}]^T$ representing the temperature, $T$, and the average shear rate, $\dot{\gamma}$, in the impeller region experienced by the fluid during mixing. For each equation $k_i$ and $K_i$ were the rate and equilibrium constants, respectively, while $H(T')$ was the Heaviside switch function for which $H(T') = 0$ for $T' < 0$ and $H(T') = 1$ for $T' \geq 0$ where $T_K$ was a key phase transition temperature crossed during the quench process.

$$r_1 = k_1 \cdot \dot{\gamma} \cdot (\alpha - T) \cdot [X_A X_W] \cdot H(T - T_K) \tag{8a}$$

$$r_2 = k_2 \cdot \dot{\gamma} \cdot T \cdot \left[ X_L X_W - \frac{X_{L^*}}{K_2 \cdot T^{-1}} \right] \tag{8b}$$

$$r_3 = k_3 \cdot \dot{\gamma} \cdot \left[ X_L - \frac{X_V}{K_3 \cdot \dot{\gamma} \cdot (T - \beta)} \right] \cdot H(T - T_K) \tag{8c}$$

The mass and energy balance, outlined in Equations 9a, 9b, 9c and 9c, was constructed for the recycle emulsification configuration shown in Figure 2a, which is common industrially. Where $V$ is volume, $Q$ is volumetric flowrate, $t$ is batch time, $T$ is temperature, and $C_p$ is the specific heat capacity. For the mixed tank heating-cooling jacket, $U$ is the specific heat-transfer coefficient, $A$ is the surface area, and $T_{LMTD}^j$ is the log-mean temperature difference. Then $\boldsymbol{X}$ was a vector of lumped phase concentrations. The superscripts $m$, $p$, $t$, and $i$ denote addition processes, while $r$ and $s$ denote the pre- and post-shear streams entering and exiting the in-line mixer, respectively. The average shear rate in an agitator region can be estimated using the relationship: $\dot{\gamma} = K_S \cdot N$, where $K_S$ is the dimensionless Metzner-Otto constant, and $N$ is the agitator speed (Metzner & Otto, 1957).

$$\frac{dV^m}{dt} = Q^i + Q^t \tag{9a}$$

$$\frac{dT^m}{dt} = \frac{C_p^t Q^t}{C_p^m V^m} T^t + \frac{C_p^s Q^s}{C_p^m V^m} T^s - \frac{C_p^m Q^b}{C_p^m V^m} T^m - \frac{T^m}{V^m} \cdot \frac{dV^m}{dt} + \frac{UAT_{LMTD}^j}{C_p^m V^m} \tag{9b}$$

$$\frac{dT^s}{dt} = \frac{Q^s}{C_p^s V^s} \cdot (C_p^r T^r - C_p^s T^s) \tag{9c}$$

$$\frac{d\boldsymbol{X}^m}{dt} = \frac{Q^t}{V^m} \cdot \boldsymbol{X}^t + \frac{Q^s}{V^m} \cdot \boldsymbol{X}^s - \frac{Q^b}{V^m} \cdot \boldsymbol{X}^m - \frac{Q^t + Q^i}{V^m} \cdot \boldsymbol{X}^m \tag{9d}$$

$$\frac{d\boldsymbol{X}^s}{dt} = \frac{Q^s}{V^s} \cdot (\boldsymbol{X}^r - \boldsymbol{X}^s) + \frac{d\boldsymbol{X}}{dt}(\boldsymbol{X}^s, \dot{\gamma}^s, T^s) \tag{9e}$$

It was assumed that shear-induced change is negligible in the low-shear environment of the mixed tank; hence, the reaction rate (i.e., $d\boldsymbol{X}/dt$) is absent from Equation 9d for the main mixer and only present in Equation 9e for the in-line mixer. The components of $d\boldsymbol{X}/dt = f(\boldsymbol{X}, \boldsymbol{C})$ are defined in Equations 10a, 10b, 10c, 10d and 10e in terms of the three reaction rates from Equations 8a, 8b and 8c, and the multi-phase product formation stoichiometry.

$$\frac{dX_A}{dt} = -2r_1(\boldsymbol{X}, \boldsymbol{C}) \tag{10a}$$

$$\frac{dX_W}{dt} = -5r_1(\boldsymbol{X}, \boldsymbol{C}) - 10r_2(\boldsymbol{X}, \boldsymbol{C}) \tag{10b}$$

$$\frac{dX_L}{dt} = +r_1(\boldsymbol{X}, \boldsymbol{C}) - r_2(\boldsymbol{X}, \boldsymbol{C}) - 3r_3(\boldsymbol{X}, \boldsymbol{C}) \tag{10c}$$

$$\frac{dX_{L^*}}{dt} = +r_2(\boldsymbol{X}, \boldsymbol{C}) \tag{10d}$$

$$\frac{dX_V}{dt} = +r_3(\boldsymbol{X}, \boldsymbol{C}) \tag{10e}$$

### 4.6. Product Formulation Process Flow Diagram

The synthetic PFD generated from the mechanistic model mirrors the four steps involved in the product's production: (i) ingredient injection, (ii) high-temperature mixing, (iii) quenching, and (iv) low-temperature mixing. For each step, there were four common processing parameters: step duration, $\tau_n$, bottom flowrate, $Q_n^b$, in-line mixer speed, $N_n^s$, and cooling-jacket flowrate, $Q_n^j$, where the subscript $n \in \{1, 2, 3, 4\}$ index the step. More step specific, there were the volumes $V_0^W$, $V_1^A$, $V_3^W$ and temperatures $T_0^W$, $T_1^A$, $T_3^W$ of specific ingredients added at the different steps. The coupled ODEs were solved using LSODA (Petzold, 1983) implemented in the Python library SciPy 1.10.1 (Virtanen et al., 2020).

### 4.7. Application of SR-MBDoE to Case Studies

In these synthetic experiments, the temperature ($T$) and lumped phase concentrations entering and exiting (i.e., $\boldsymbol{X}^r$ and $\boldsymbol{X}^s$, respectively) the in-line mixer were measured each minute throughout the batch, while the in-line mixer speed (i.e., $N^s \propto \dot{\gamma}^s$) was controlled. The most influential processing parameters to be included in the experiment parameter design space ($\boldsymbol{\vartheta}$) were the initial temperature, $T_0$, the step duration, $\tau_n$ for $n \in \{2, 3, 4\}$, and the in-line mixer speed, $N_n$ for $n \in \{2, 4\}$, where $n \in \{1, 2, 3, 4\}$ indexes the processing step number. Their lower and upper bounds (i.e., $\boldsymbol{\vartheta}_{lb}$ and $\boldsymbol{\vartheta}_{ub}$) were set based on their typical ranges from experience. All other processing parameters were fixed to their nominal values.

Provided that the residence time inside the in-line mixer is small, the rearrangement rates (i.e., $\mathrm{d}\boldsymbol{X}/\mathrm{d}t$) can be approximated using Equation 11. Hence, the left-hand-side of the Equations 10a, 10b, 10c, 10d and 10e were known allowing the reaction rates (i.e., $r_1$, $r_2$ and $r_3$) to be found by solving the resulting set of simultaneous algebraic equations. With these reaction rates, SR then aims to recover Equations 8a, 8b and 8c as functions of the lumped phase concentrations (i.e., $X_W$, $X_A$, $X_L$, $X_V$ and $X_{L^*}$) and operating conditions (i.e., $T$ and $\dot{\gamma} \propto N$). The proposed rate expressions could later be substituted back into Equations 10a, 10b, 10c, 10d and 10e, and those substituted into the mass and energy balance in Equation 9d to simulate the PFD.

$$\frac{\mathrm{d}\boldsymbol{X}}{\mathrm{d}t} \approx \frac{\Delta \boldsymbol{X}}{\Delta t} = \frac{\boldsymbol{X}^s - \boldsymbol{X}^r}{V^s/Q^s} \tag{11}$$

Expressions were searched up to a complexity (i.e., the total number of operators, variables, and constants) of twenty. The five lumped phase concentrations and operating conditions were considered, as well as real-valued constants. The operators considered were addition, subtraction, multiplication, and division. To minimise the risk of entrapment in a suboptimal local solution, 45 populations were evolved in parallel until a maximum of 10,000 iterations of *evolve-simplify-optimise* or the MSE converged. Input and output features were all rescaled before SR by division by their maximum absolute values.

To initiate MBDoE, four synthetic experiments were conducted at high and low initial temperatures, $T_0^W$, and step two in-line mixer speeds, $N_2^s$, yielding the product KPI profiles seen earlier in Figure 4b. Then, for each iteration, SR proposed nine possible expressions approximating the rates (i.e., three for each $r_1$, $r_2$ and $r_3$ rate expression) as functions of the lumped phase concentrations (i.e., $X_W$, $X_A$, $X_L$, $X_V$ and $X_{L^*}$) and operating conditions (i.e., $T$ and $\dot{\gamma} \propto N$). Next, one new experiment minimising $J(\vartheta)$ was proposed and conducted, expanding the dataset for the next iteration. MBDoE was repeated for ten iterations or until the ground truth was recovered. Note that although the system is not explicitly a function of step duration (i.e., $\tau_n$), each combination of processing parameters nonetheless yields a unique process trajectory and new information.

### 4.9. Prediction Accuracy

For each rate equation, three possible candidates of different complexities were proposed – these were used to estimate the variance and guide MBDoE towards the operating conditions with the greatest variance – but to make a point prediction of the final product KPI, only the single top-scoring expression for each rate equation was used. This point prediction of the product KPI was validated against a large set of random in-silico experiments not used in model construction, and the mean absolute percentage error (MAPE) was computed.

## S1. Supplementary Results and Discussion

### S1.1. Experiment for Model Discrimination

In general, $r_1$ was the easiest to recover, then $r_3$, while $r_2$ was more difficult, but all three must be well approximated for the MAPE to be small. Tables S1 and S2 show the top three scoring expressions at $I_{MBDoE} = 1$ and $I_{MBDoE} = 2$, respectively, for Case Study 1A. In $I_{MBDoE} = 1$ the error is from $r_2$ (i.e., responsible for $L + 10W \rightleftharpoons L^*$) with three mistakes: (i) $N^s$ is absent from the top scoring expression; and (ii) T and (iii) $X_L$ are found in the denominator of the reverse driving force term since both are high early in the batch. The other lower-scoring candidates do include $N^s$, albeit incorrectly and place $X_L$ differently or $X_{L^*}^{-2}$ in the denominator. The first iteration of MBDoE aimed to discriminate between these expressions by proposing an experiment with operating conditions that maximised the variance, $-J_M(\vartheta)$, in the final KPI. To visualise the methodology and the proposed experiment, Figures S1a and S1c show the propagated bounds generated by combinations of the expressions in Table S1 for three different experiments (i.e., green, red and blue). The blue KPI trajectories are those generated under the operating conditions that maximise the variance and represent the experiment proposed by MBDoE for $I_{MBDoE} = 2$. The green and red KPI trajectories exemplify how poor choice in operating conditions makes harder model discrimination. Figures S1b and S1c show two slices through the five-dimensional objective $J_M(\vartheta)$ about the minimum (i.e., the blue point). It can be seen how it was proposed to conduct a long $\tau_2 = 60$ min step two at high-temperature at low-shear, followed by a quick $\tau_3 = 5$ min quench. Low shear at high temperature minimised $X_L$ so that T and $X_L$ were no longer so strongly correlated; meanwhile $N^s$ could no longer be approximated as constant, manifested in the larger variance in Figure S1b at high temperatures away from the centre line at an intermediate shear. The proposed experiment makes sense from a model discrimination perspective and allowed $r_2$ to be correctly identified in $I_{MBDoE} = 2$.

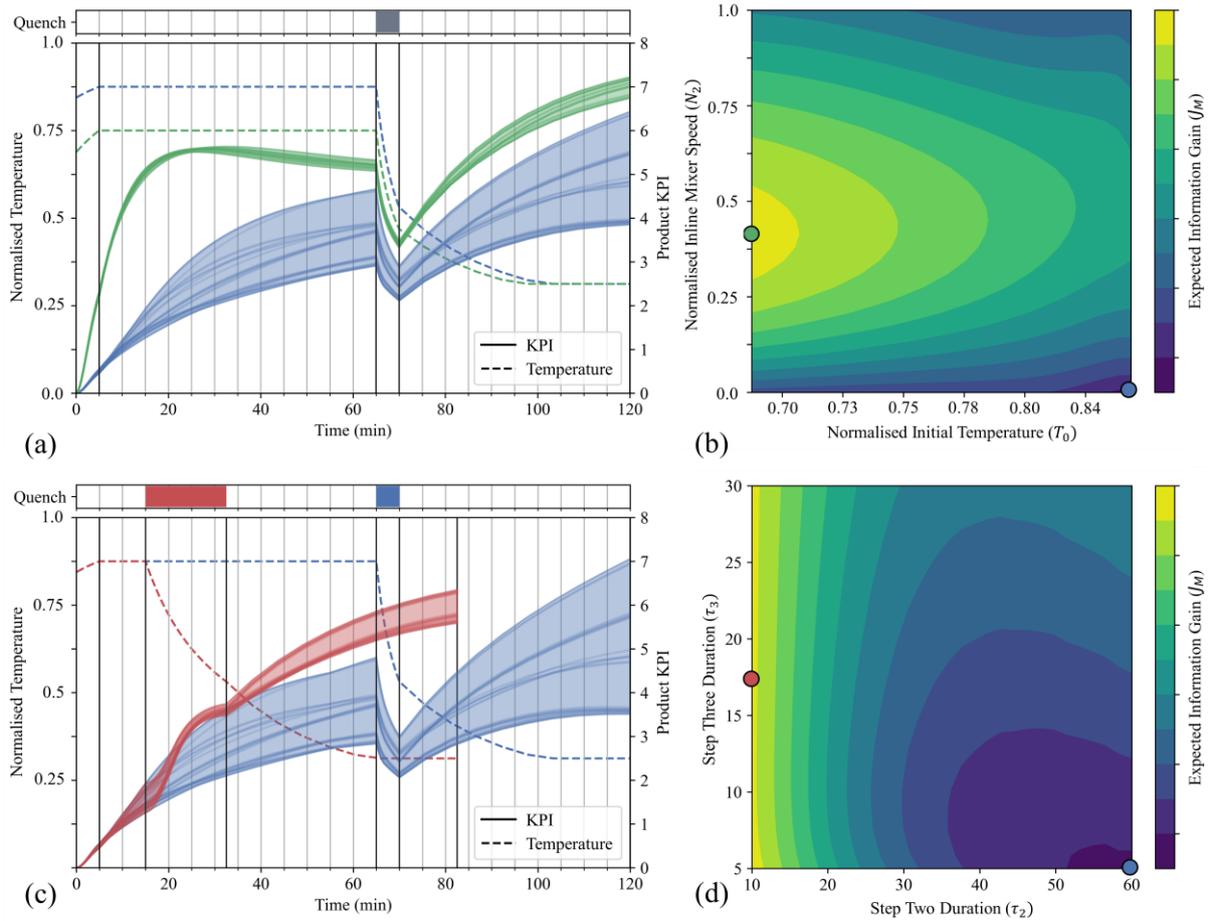

Figure S1: Simulated KPI bounds (left) generated by combinations of top-scoring expressions at $I_{MBDoE} = 1$ for Case Study 1A at different $T_0^W$ and $N_2^s$ (top) and different $\tau_2$ and $\tau_3$ (bottom). Two slices of $J(\vartheta)$ through the new optimal experiment (right), represented by the blue point that maximised the variance. The green and red points and corresponding KPI trajectories exemplify poor experimental design that instead minimised variance.

Table S1: Top three scoring expressions for each rate equation (i.e., $r_1$, $r_2$ and $r_3$) for Case Study 1A for $I_{MBDoE} = 1$ following the addition of a new experiment. Note that complexity was evaluated before simplification and substitution of normalisation coefficients.

| Complexity | MSE | Score | Equation |
|---|---|---|---|
| 13 | 4.9E-13 | 10.97 | $r_1 = 0.00013 N T X_A X_W (-0.79 + 56.0/T)$ |
| 11 | 1.7E-03 | 1.15 | $r_1 = 0.003 N X_A X_W (-1.9 + 140.0/T)$ |
| 9 | 1.6E-02 | 1.04 | $r_1 = 3.0 X_A X_W (-4.3 + 310.0/T)$ |
| 11 | 3.3E-02 | 0.38 | $r_2 = 0.013 X_L \left( -\dfrac{150.0 X_{L^*}}{-0.042T + 63.0 X_L + 3.0} + 3.3 X_W \right)$ |
| 15 | 1.4E-02 | 0.23 | $r_2 = 0.013 X_L \left( -\dfrac{0.00067 T}{X_L (4.2 \times 10^{-8} \cdot M / X_{L^*}^2 + 1.0)} + 3.3 X_W \right)$ |
| 13 | 2.3E-02 | 0.19 | $r_2 = 0.013 X_L \left( -\dfrac{0.1 T X_{L^*}}{X_L (0.001 N + 1.2)} + 3.3 X_W \right)$ |
| 17 | 7.1E-05 | 1.61 | $r_3 = 6.4 \times 10^{-6} \cdot N \left( 63.0 X_L - \dfrac{140000 X_V}{N(0.042T - 2.2)} \right)$ |
| 11 | 1.2E-02 | 1.12 | $r_3 = 0.96 X_L - \dfrac{2700 X_V}{N(0.042T - 2.1)}$ |
| 15 | 1.8E-03 | 0.51 | $r_3 = 6.0 \times 10^{-6} \cdot N \left( 63.0 X_L - \dfrac{140000 X_V}{N(0.042T - 2.2)} \right)$ |

Table S2: Top three scoring expressions for each rate equation (i.e., $r_1$, $r_2$ and $r_3$) for Case Study 1A for $I_{MBDoE} = 2$ following the addition of a new experiment. Note that complexity was evaluated before simplification and substitution of normalisation coefficients.

| Complexity | MSE | Score | Equation |
|---|---|---|---|
| 13 | 7.0E-13 | 11.85 | $r_1 = 0.00013 N T X_A X_W (-0.79 + 56.0/T)$ |
| 9 | 1.5E-02 | 0.97 | $r_1 = 3.0 X_A X_W (-4.4 + 310.0/T)$ |
| 11 | 1.4E-02 | 0.05 | $r_1 = 0.13 T X_A X_W (-1.8 + 130.0/T)$ |
| 17 | 6.9E-10 | 8.52 | $r_2 = 6.2 \times 10^{-6} \cdot T N X_L (-0.033 T X_{L^*}/X_L + 3.3 X_W)$ |
| 19 | 1.9E-12 | 2.95 | $r_2 = 6.2 \times 10^{-6} \cdot T N X_L (-0.033 T X_{L^*}/X_L + 3.3 X_W)$ |
| 11 | 3.3E-02 | 0.35 | $r_2 = 0.013 X_L \left( -\dfrac{2.4 X_{L^*}}{X_L (3.8 - 0.042 T)} + 3.3 X_W \right)$ |
| 15 | 3.4E-04 | 1.51 | $r_3 = 0.0004 N X_L - \dfrac{1.2 X_V}{0.054 T - 2.8}$ |
| 19 | 2.8E-06 | 1.28 | $r_3 = 4.7 \times 10^{-6} \cdot N \left( 85.0 X_L - \dfrac{180000 X_V}{N(0.042 T - 2.2)} \right)$ |
| 13 | 7.0E-03 | 1.27 | $r_3 = 0.00042 N X_L - \dfrac{1.2 X_V}{0.042 T - 2.1}$ |

## S1.2. Physical Interpretation of Optimum PFD

Table S3 shows the optimised PFD processing parameters. For this synthetic system, the formation of $X_{L^*}$ phase is responsible for KPI build. To achieve a high KPI, there is a trade-off between step two and three: a short step two (i.e., $2A + 5W \rightarrow L$) will stunt long-term $X_{L^*}$ formation capacity (i.e., $L + 10W \rightleftharpoons L^*$) but a long step two creates more $X_V$ phase (i.e., $3L \rightleftharpoons V$) indirectly subtracting from the amount of $X_{L^*}$ phase. By examining the optimised PFD processing parameters, it is found that it was optimal to minimise $T_0^W$ and $N_2^s$ to minimise $X_V$ phase formation in step two, proceeding only up to the point at which the rate of $X_V$ phase formation became significant. While a slow quench could improve product KPI build by reversing some $X_V$ phase to $X_L$ phase which can become $X_{L^*}$ phase, a quick quench saves time, and then a high in-line mixer speed in step four accelerated the final KPI build to specifications.

Table S3: Optimised PFD processing parameters ($\vartheta$) at the end of MBDoE of Case Study 2. $T_0$ is the initial temperature, $\tau_n$ is the step duration, $N_n$ is the in-line mixer speed, while the subscript $n \in \{1, 2, 3, 4\}$ indexes the processing step number. All units are normalised between zero and one between their lower and upper bounds in experiment design space.

| Processing Parameter | Normalised Value |
| --- | --- |
| $T_0$ | 0 |
| $\tau_2$ | 0.11 |
| $N_2$ | 0 |
| $\tau_3$ | 0 |
| $N_4$ | 0 |
| $\tau_4$ | 0.025 |